\documentclass[runningheads]{llncs}

\usepackage{geometry}
\usepackage[T1]{fontenc}
\usepackage{lmodern}
\usepackage{graphicx}
\usepackage{cite}
\usepackage{amsmath,amssymb,amsfonts}
\usepackage{algorithm}
\usepackage{algorithmicx}
\usepackage{algpseudocode}

\usepackage{textcomp}
\usepackage{xcolor}

\providecommand{\doi}[1]{doi: {\footnotesize \href{http://dx.doi.org/#1}{\path{#1}}}}

\usepackage[pdftex=true,breaklinks=true,hidelinks=true,colorlinks=true,citecolor=blue]{hyperref}

\usepackage{graphicx}
\graphicspath{{pdf/}{photo/}{figures/}}
\DeclareGraphicsExtensions{.eps,.jpeg,.jpg,.png,.pdf}

\usepackage{subcaption}
\usepackage[]{caption}

\usepackage{fancyhdr}
\usepackage{booktabs}

\def\BibTeX{{\rm B\kern-.05em{\sc i\kern-.025em b}\kern-.08em
    T\kern-.1667em\lower.7ex\hbox{E}\kern-.125emX}}

\usepackage{orcidlink}

\begin{document}
\title{Multi-Agent Object Detection Framework Based on Raspberry~Pi YOLO Detector and Slack-Ollama Natural Language Interface}

%
%
\author{Vladimir Kalušev\inst{1}\orcidlink{0009-0005-8851-5790} \and
Branko Brkljač\inst{2}\orcidlink{0000-0001-7932-6676} \and
Milan Brkljač\inst{3}\orcidlink{0000-0002-0617-973X}}
\authorrunning{V. Kalušev, B.Brkljač and M.Brkljač}
%
\institute{The Institute for Artificial Intelligence Research and Development of Serbia \\
Fruškogorska 1, 21000 Novi Sad, Republic of Serbia \\
\email{vladimir.kalusev@ivi.ac.rs}
\and
Department of Power, Electronic and Telecommunication Engineering \\
Faculty of Technical Sciences, University of Novi Sad\\
Trg Dositeja Obradovića 6, 21000 Novi Sad, Republic of Serbia \\
\email{brkljacb@uns.ac.rs} \and
Faculty of Finance, Banking and Auditing, Alfa BK University \\
Bulevar maršala Tolbuhina 8, 11070 Belgrade, Republic of Serbia \\
\email{milan.brkljac@alfa.edu.rs}
}
\maketitle              
\begin{abstract}
\setcounter{footnote}{0}
The paper presents design and prototype implementation of an edge based object detection system within the new paradigm of AI agents orchestration. It goes beyond traditional design approaches by leveraging on LLM based natural language interface for system control and communication and practically demonstrates integration of all system components into a single resource constrained hardware platform. The method is based on the proposed multi-agent object detection framework which tightly integrates different AI agents within the same task of providing object detection and tracking capabilities. The proposed design principles highlight the fast prototyping approach that is characteristic for transformational potential of generative AI systems, which are applied during both development and implementation stages. Instead of specialized communication and control interface, the system is made by using Slack channel chatbot agent and accompanying Ollama LLM reporting agent, which are both run locally on the same Raspberry Pi platform, alongside the dedicated YOLO based computer vision agent performing real time object detection and tracking. Agent orchestration is implemented through a specially designed event based message exchange subsystem, which represents an alternative to completely autonomous agent orchestration and control characteristic for contemporary LLM based frameworks like the recently proposed OpenClaw. Conducted experimental investigation provides valuable insights into limitations of the low cost testbed platforms in the design of completely centralized multi-agent AI systems. The paper also discusses comparative differences between presented approach and the solution that would require additional cloud based external resources. Since the new AI paradigms are also affecting the current labor market, we are also briefly discussing possible implications of the ongoing changes on the way how the existing products and services are being made.

\end{abstract}

\section{Introduction}
\begin{figure}[htb]
  \centering
  \begin{subfigure}{0.7\linewidth}
    \centering
    \includegraphics[width=\linewidth]{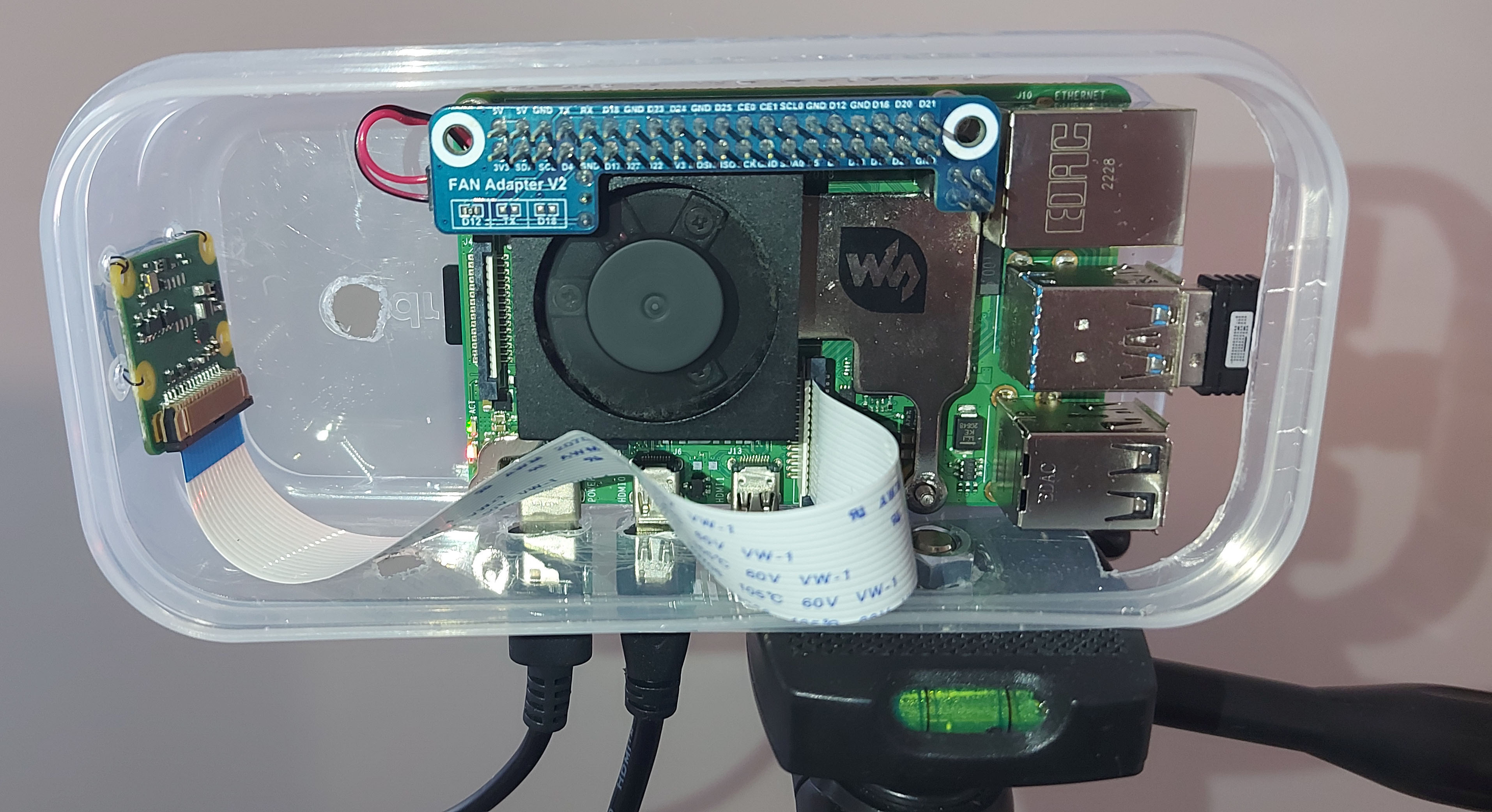}
    \caption{}
    \label{fig: Hardware testbed_a}
  \end{subfigure}
  \begin{subfigure}{0.29\linewidth}
    \centering
    \includegraphics[width=\linewidth]{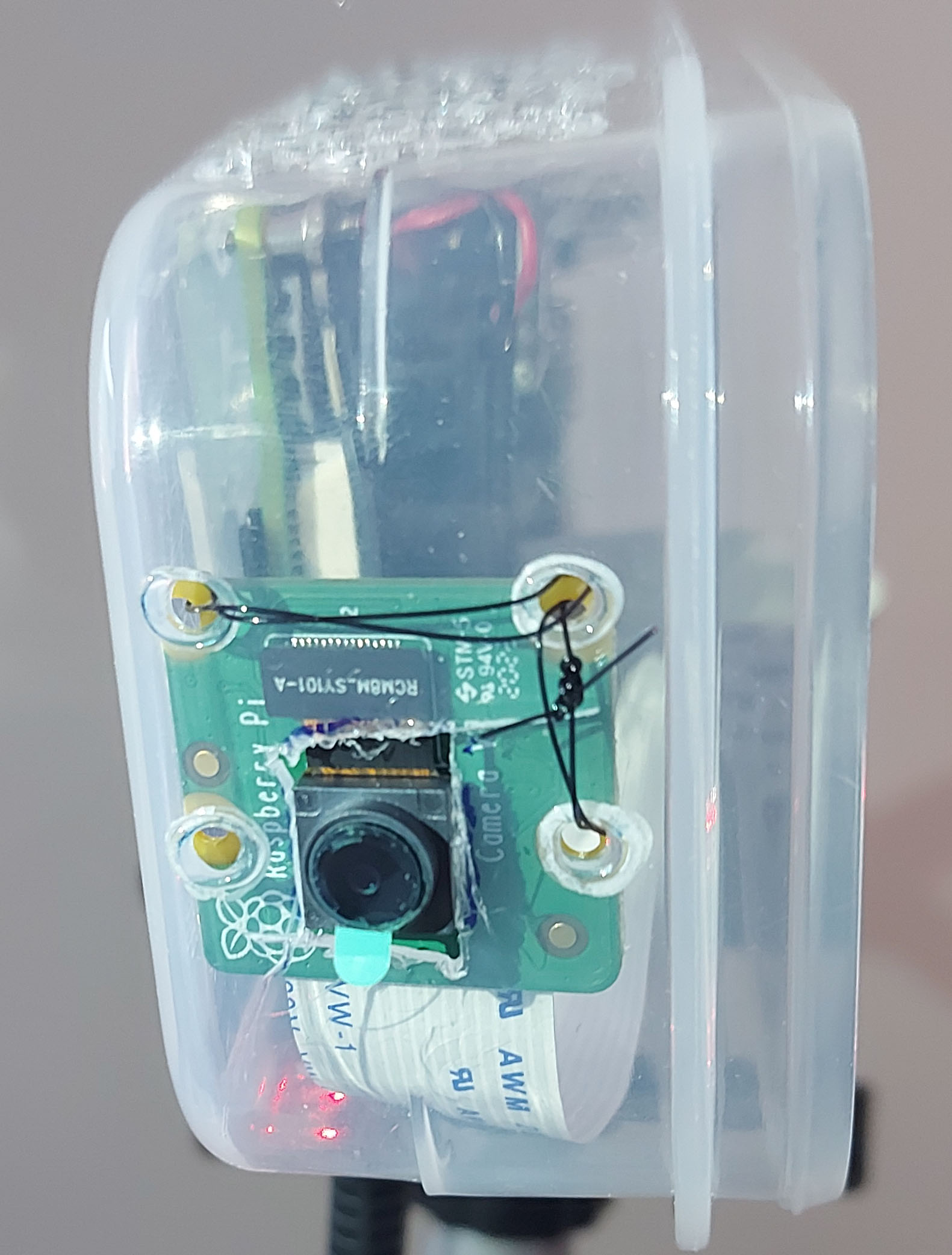}
    \caption{}
    \label{fig: Hardware testbed_b}
  \end{subfigure}
  \captionsetup{justification=justified}
  \caption{Developed fast prototyping hardware testbed: (a)~custom Raspberry Pi camera setup, (b)~MIPI CSI camera board with Sony IMX219 color sensor.}
  \label{fig: Hardware testbed}
\end{figure}

Recent years have witnessed significant advancements in the fields of computer vision and natural language processing. Proliferation of large language models (LLMs) and AI enabled edge devices have brought novel opportunities for integration of different functionalities and improved system performance. In particular, the traditional product development paradigm based on costly and error prone steps with slow incremental improvements is consistently being replaced by fast prototyping characterized by multiple trial and error development cycles and significantly lower cost per iteration. In such fast pace market environment the key to success is often based on efficient exploitation of novel capabilities brought by generative AI systems and their transformational potential \cite{GenAI}. What were previously regarded as separate tasks and required dedicated teams and resources can nowadays be accomplished with fraction of previous costs by specialized software development frameworks and generic low-cost hardware implementation platforms. In the context of such trends, which are contributing to higher proliferation of AI technologies and fostering creativity and innovation in varying application scenarios across different industries, the paper investigates transformational potential of Generative AI in development and operation of multi-agent vision based systems. The study showcases design and implementation of an edge based prototype with real-time processing capabilities and high level integration of several subsystems implementing different AI agent functionalities. The main goal of the study was to demonstrate that system integration of edge based components capable of performing complex vision tasks can be performed without costly development of application specific control and communication subsystems \cite{luu2022vei}, but instead based on contemporary multi-agent AI paradigm and utilization of natural language interface that will play the same role. The hypothesis was that such integration could be performed in fully centralized way, in which all AI agents and their orchestration would be run on standardized, commercially available low-cost hardware, without the need for additional external resources. In order to demonstrate such generic design goals we have selected object detection \cite{zhao2019object} and tracking \cite{wen2020ua}, as some of the most common vision tasks, and Raspberry Pi \cite{RaspberryPi4specs2026} as an example of ubiquitously available AI enabled edge platform on which such approach could be fully tested. Developed fast prototyping hardware testbed with custom camera integration and fan equipped device casing is illustrated in Fig.~\ref{fig: Hardware testbed}.

 \vspace{1em}

The proposed multi-agent object detection framework explores design approach in which AI agents are actively collaborating and competing for shared resources on the same platform, but are not fully autonomous nor autonomously built from scratch, which would be characteristic for some of the most recent and alternative design approaches. Namely, in comparison to traditional AI agent orchestration in which their deployment, collaboration and communication are based on adaptive but in essence still, to the large extent, deterministic programs, some of the most recent trends consider the AI agents orchestration from the more open-ended and non-deterministic design perspectives. The best example is the recently announced OpenClaw \cite{OpenClawAI} orchestration framework, in which (besides the orchestration task) automated LLM inference and host system resources can be also used to build a platform specific multi-agent AI systems from scratch. Main characteristics of such non-traditional design approaches is that the complete system design and agent architectures can be described and implemented by using only natural language, as well as their orchestration.

Motivated by the recent success of such fully autonomous and high level AI automation tools, we have decided to explore an alternative approach that could be feasible to implement on resource constrained hardware platforms in the described fully centralized way. Although some of the first considerations also included system design based on the described OpenClaw paradigm, after initial tests it was concluded that such approach could not be feasible due to limited performance of locally run LLM and high demand for LLM based inference at both development and deployment phases of such multi-agent systems. Instead, it was decided to pursue a hybrid solution that would incorporate the best of the both worlds, natural language interface for smooth and efficient communication with multi-agent system, and an event based message exchange subsystem for task specific orchestration of locally run AI agents. In the case of practical applications that require complex system components that can allocate a significant part of available development budget and resources, e.g. development of specific vision based analytics subsystems, the advantages of utilizing LLM based natural language interface for system control and communication are that the costs of wider system integration of such subsystems with costly development can be significantly lowered by leveraging on the existing generative AI technologies.

Thus, the presented object detection framework is proposed as the generic template for fast prototyping of similar fully centralized multi-agent AI systems on resource constrained platforms, enabling complete multi-agent system integration and testing in short development cycles. We note that the use of generative AI can be present at both development and operational level of such systems. In the proposed framework we assume that generative AI in the form of locally run LLM is utilized in automating system operation, while the automated generation of system software components is not pursued due to high computational cost of locally run LLM inference. However, this does not exclude  alternative development scenarios in which the development process would be performed externally and independently from the selected deployment platform, e.g. by utilizing externally run LLM inference or complete platform emulation, allowing for full development automation even in the case of the proposed hybrid design solution presented in this study.

The rest of the paper is organized as follows. In Section 2 we provide an overview of the related work, with particular emphasis on computer vision tasks of object detection and tracking, natural language interfaces based on open source LLMs, automated chatbots, and multi-agent AI systems. In Section 3 we describe the design of the proposed multi-agent object detection framework and practical implementation details of the developed AI agents, which are run locally and performing vision, reporting, and communication and control tasks on the single Raspberry Pi device. Results of real-time system operation consisting of YOLO type object detection, passive tracking and interaction with the Slack \cite{Slack2026} messaging application chatbot running on the locally deployed Ollama \cite{Ollama2026} LLM are presented in Section 4. It also includes additional discussion of the low-cost multi-agentic framework design, its comparison to the OpenClaw design principles, and possible market implications of the generative AI based fast prototyping. Finally, in Section 5 we refer to main conclusions and possible future research directions.

\section{Related work}

Integration of different AI subsystems into single embedded computing platform has gained significant momentum. From home assistants, smart buildings, conversation systems, home automation, to surveillance systems, traffic monitoring, process control, autonomous driving, medicine, or the augmented reality and the gaming industry, the need for combining different aspects of AI is becoming ever more important across range of different industries. Reasons lie in the need for achieving better functionality of the corresponding products and services, but also in providing higher flexibility in system development and deployment, with both aspects resulting in easier adaptation of generic AI solutions to wider range of the user specific application scenarios. Thus, on the one hand there is a significant progress at the level of hardware miniaturization and its software abstraction, which is characteristic for numerous currently available commercial development boards or their open source counterparts. While, on the other hand, the ongoing algorithmic improvements in the fields of computer vision (CV) and natural language processing (NLP) are opening numerous opportunities for ideation and prototyping of innovative technical systems that are exploiting high level inference capabilities of complex AI models \cite{brkljavc2025applications}. In such environment the next logical step is a convergence of these technologies towards a unified and diverse technical ecosystem in which AI systems will be easier to bring to the market and test under real operational conditions and user acceptance levels. In this process the role of natural language interfaces will seem to be twofold. Primarily they will provide natural and easier communication with human users, without the need for specialized training or technical knowledge, lowering the barrier for complex technology acceptance. However, natural language interfaces will be also the best bridge towards automated development of agentic AI systems, as a universal language for providing necessary instructions and constraints for the software development process ideation, as well its automation at various stages of continuous integration,  delivery and deployment (CI/CD). The second claim is based on the already existing trend of increased usage of generative AI tools in the software engineering industry, which are reducing the development costs and increasing work productivity of the employees. In that sense, natural language interfaces can be at the core of development process, or have a more traditional role in providing easier command and control of other AI subsystems. In this work, we are exploring the latter one, by demonstrating design of an agentic AI system in which  Ollama LLM is combined with Slack channel chatbot application in order to provide functional communication and control in fast prototyping of complex vision system on an embedded device.

Object detection is one of the most explored high level vision tasks with strong research track record. Despite its maturity there are still various types of challenges like occlusions, clutter, adverse operating conditions or  complex and dynamic scenes, affecting the overall speed and accuracy of the currently available algorithms.  Thus, there  are various design improvements that are constantly being made in order to improve algorithm performance and robustness. Notable examples include combining multiple sources of information, addressing multi-scale nature of scene objects at both image and feature levels, applying highly complex hierarchical feature extraction mechanisms based on deep neural networks \cite{pal2021deep}, adaptive attention mechanisms at various levels of information extraction process, or optimizing the postprocessing steps and final classification and regression operations, which are key for successful object detection and its precise localization.

Historically, the object detection paradigm has gone through the gradual transformation from sliding window approaches and the two stage detectors \cite{girshick2015region, ren2016faster} towards the end-to-end predictions of object location and type \cite{yolo}. In that process the recent change of focus from convolutional neural networks (CNNs) towards attention based mechanisms and transformer based architectures that are capable of capturing the global context of the object detection task have contributed to high variety of possible architectures and design choices. Although transformer based architectures like DETR \cite{DETR} have been proven to be on pair with computationally efficient YOLO architectures, the wider engineering community still seems to be preferring the well established CNN architectures and their improved variants. From production side point of view, specialized software development frameworks offering high quality of the shelf implementations of the single stage detectors from the YOLO family have been widely accepted by the software development community due to their constant maintenance, benchmarking and possibly easier porting to specific hardware platforms, which are usually from the start designed to fully support their software stack.

As the main algorithmic solution for the demonstration of the proposed multi-agent  object detection framework we have decided to rely on the YOLO implementations provided by the REF. and specifically refer to some of the less demanding variants like Yolov8n. We note that there could also be numerous other solutions, depending on available hardware platform and design requirements. However, the selected choice was made in particular to demonstrate the fast prototyping on the available Raspberry Pi platform, which has evolved from general purpose ARM based system on chip (SoC) to high quality AI enabled edge computing device. When it comes to LLMs as one of the key enabling technologies for the design of natural language interfaces, we have relied on Ollama open-source tool for running LLMs locally on the selected Raspberry Pi device. Multi-agent systems can be based on diverse system architectures and agent types, but their design is usually driven by the desired system operation and infrastructure requirements. Thus, we have followed and adopted an approach in which the proposed solution should be easy to adapt to different resource constrained platforms and support diverse agent configurations and their interactions in order to solve complex tasks.

\section{Design and Development}
Object detection \cite{zou2023object} and multi-agent systems are well known concepts that have been systematically explored in different contexts. The key research question of this study was to investigate feasibility of implementing a combination of these complex functional concepts on the edge device like Raspberry Pi platform. It was done in order to test the hypothesis that all AI agents and their orchestration can be run locally on the low-cost hardware, without the need for additional external resources like LLM APIs or other processing off-loading solutions. In addition, in order to demonstrate the current trend of exploiting generative AI tools for cutting costs and improving productivity in the software engineering industry, i.e. providing fast prototyping capabilities, we have decided to demonstrate the use of such tools in development of high fidelity natural language interface, without the need for full development of specialized communication and control for designed multi-agent system. Namely, an additional advantage of using user familiar environment like the Slack channel chatbot for receiving system generated messages and reports, or giving commands to control agent and system orchestrator is that the potential users or testers of the deployed system do not need to be technically knowledgeable, nor go through time consuming training for system operation. Instead, the feedback can be received almost instantly on a relatively large sample of potential system testers. Development and implementation of the showcased prototype have also been done as a design template for similar application scenarios or multi-agent systems that would run under same conditions on the low-cost hardware. Thus the focus of the proposed object detection framework was on the practical realization of the multi-agent system and investigation of its limitations.\footnote{Implementation code will be made available upon manuscript publication.}

\begin{figure}[htb]
    \centering
    \includegraphics[width=\linewidth]{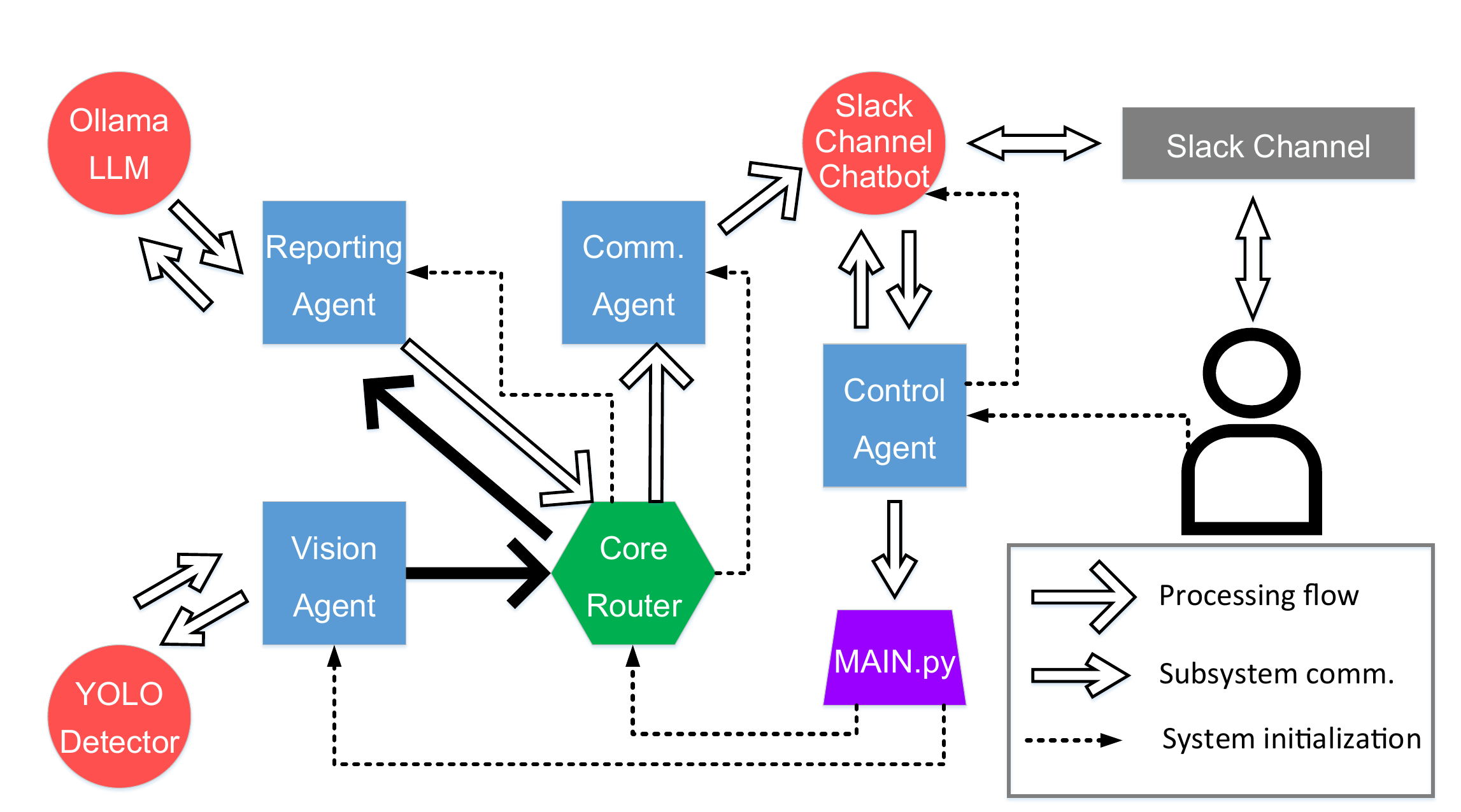}
  \captionsetup{justification=justified}
  \caption{Multi-agent system architecture: functional elements of the proposed object detection framework and implemented agentic components. }
  \label{fig: System architecture}
\end{figure}

\subsection{Multi-Agent System Architecture}

The proposed system consists of several agentic components that are loaded into memory and continuously providing different types of functionalities. Their orchestration is based on an event driven pipeline, which goes through specifically designed message router or core message passing interface. The router is facilitating different event callbacks and coordinating execution of designed agentic components. In essence, it is traditional event driven framework, but with the exception of being run on the low-cost hardware and with AI components that in general case can have non-deterministic behavior. Since it was designed for the relatively low resource embedded platform, agent orchestration was not done by a specialized AI agent or AI based middleware, but instead by the developed low complexity router that accepts predefined set of agents and agent generated events. Thus, instead of running an AI based orchestrator, e.g. like OpenClaw, system resources have been mostly utilized for specific AI subsystems of designed agentic components. This is illustrated in Fig.~\ref{fig: System architecture}, where the relationships between vision, reporting, communication and control agents and their orchestration through the designed core router are depicted, alongside their AI subsystems and user interaction.

 It can be seen that the main information or processing flow during system operation goes from the vision agent, as the source of environment perception based on the real-time camera feed analysis.  Vision agent asynchronously propagates event messages via system core or router component towards LLM equipped reporting agent, which compiles event triggered natural language reports. Reports are then sent again via router towards communication agent, which is waiting for reporting events and operates preloaded chatbot. After receiving a message, it forwards information in the form of text and images to the corresponding chat channel, which can be also accessed by the subscribed user. Thus, the user receives information  through the familiar mobile phone or web client interface, and if necessary sends commands to the control agent that is subscribed to the same chat channel. It is also the main way of system initialization, which is performed by sending a message to the preloaded control agent, which then starts or stops the remaining system components. This process is also illustrated in Fig.~\ref{fig: System architecture} by the corresponding dashed arrows, while the long arrows depict the described event triggered information flow during system operation. Short arrows are representing local communication between agents and their AI enabled subsystems.

 Multi-agent setup makes clear distinction between responsibilities and scopes of different modules or subsystems, making it highly modular and easy to upgrade and maintain on the long run. It also allows flexibility and modularity in system design by supporting additional agents integration or system independent bridges towards external services or systems that would rely on the information generated by the multi-agent system. In the presented object detection use case these additional services could be email reporting, database filling, processing of multiple camera feeds, GPS locations, map browsing, etc. It is also highly scalable, in the sense that the choice of generic and ubiquitously available hardware could guarantee sustained production of system according to market needs. On the technical side, adding more resources to the selected platform \cite{papakyriakou2023high} would directly scale with the available agentic capabilities. Real-time processing and coordination of multiple components, i.e.  successful integration of heterogeneous AI models from CV and NLP domains makes the proposed system architecture also widely applicable to different tasks, beyond demonstrated object detection.

\subsection{Resource Constrained Implementation}

Described engineering task is constrained by the relatively low memory and computational resources of the selected Raspberry Pi System on Chip (SoC) platform, Fig.~\ref{fig: Hardware testbed},  as well as by the imposed requirements of having multiple AI components running at the same time on the local machine.  Specific hardware characteristics  \cite{RaspberryPi4specs2026} are reviewed in Table~\ref{tab: Platform_characteristics} and indicate parallel processing capabilities, but without neural processing unit (NPU) or other neural networks hardware acceleration capabilities. Nevertheless, reported peak performance of 9.96 Gflops  \cite{papakyriakou2023high} combined with hardware accelerated H.264 video encoding makes the platform suitable for adequately designed edge AI workloads.

\begin{table}[tb]
\caption{Hardware testbed specifications.}
\label{tab: Platform_characteristics}
\centering
\begin{tabular}{p{3.5cm} p{8.5cm}}
\toprule
{Component} & {Details, Raspberry Pi 4 model B} \\
\midrule
Processor & Broadcom BCM2711, Quad core Cortex-A72 (ARM v8) 64-bit SoC @ 1.8GHz \\
Memory & 8 GB LPDDR4-3200 SDRAM \\
Wireless & 2.4 GHz and 5.0 GHz IEEE 802.11ac wireless, Bluetooth 5.0 \\
Ethernet & Gigabit Ethernet \\
USB ports & 2 USB 3.0 ports; 2 USB 2.0 ports \\
GPIO & Raspberry Pi standard 40 pin GPIO header \\
Display & 2 × micro-HDMI ports (up to 4Kp60) \\
Camera/Display ports & 2-lane MIPI DSI display port; 2-lane MIPI CSI camera port \\
Audio/Video & 4-pole stereo audio and composite video port \\
Hardware codec &  H264 (1080p30 encoder), H.264 and H.265 decoder \\
Graphics & OpenGL ES 3.1, Vulkan 1.0 \\
Storage & Micro-SD card slot for OS and data storage \\
Power & 5V DC via USB-C or GPIO (min 3A); PoE (separate HAT) \\
Temperature & 0 – 50°C ambient \\
\bottomrule
\end{tabular}
\end{table}

As already mentioned in the introduction, generic design goal was to perform AI subsystems integration in a fully centralized way without the need for additional external resources for their operation.  Therefore, software implementation of the architecture depicted in Fig.~\ref{fig: System architecture} followed a minimal viable or lean architecture approach, by implementing only the main agentic components and subsystems needed for the proposed object detection framework. It avoids building dedicated web dashboards or communication protocols, but instead focuses on repurposing of the existing infrastructure like Slack's messaging for both commands and status updates based on natural language interface. This provides more resources for computationally heavy tasks like real-time object detection and tracking, and LLM based reporting. However, it also preserves necessary functionality for fast prototyping and testing of developed solution by leveraging on Slack communication channel authentication, persistance, multiple user notifications or nested conversation chains.

\begin{table}[htb]
\caption{Camera board specifications.}
\label{tab: Camera_characteristics}
\centering
\begin{tabular}{p{4cm} p{8cm}}
\toprule
{Specification} & {Details, Raspberry Pi module v2} \\
\midrule
Resolution & 8 megapixels \\
Max Video & 1080p47 \\
Still Modes & 1640 × 1232p41; 640 × 480p206 \\
Sensor & Sony IMX219 \\
Sensor Resolution & 3280 × 2464 pixels \\
\bottomrule
\end{tabular}
\end{table}

From  Fig.~\ref{fig: Hardware testbed_b} it can be seen that the utilized camera board with IMX219 sensor, Table~\ref{tab: Camera_characteristics}, is connected through MIPI serial interface, while the CPU cooling of the prototype is enhanced by an additional fan adapter \cite{WavesharePiFan2026}, Fig.~\ref{fig: Hardware testbed_a}. Developed testbed is running  Debian Linux distribution \cite{RaspberryPiOS2026} and is also equipped with HDMI to USB video capture card for recording of system operation.

\begin{figure}[tb]
  \centering
  \begin{subfigure}{0.35\linewidth}
    \centering
    \includegraphics[width=0.7\linewidth]{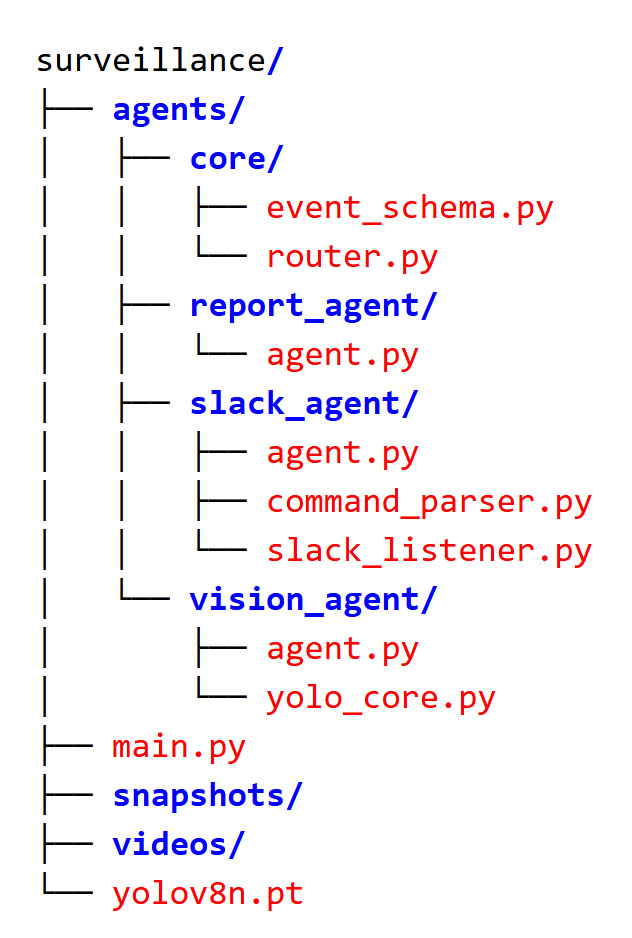}
    \caption{}
    \label{fig: Software components_a}
  \end{subfigure}
  \begin{subfigure}{0.64\linewidth}
    \centering
    \includegraphics[width=\linewidth]{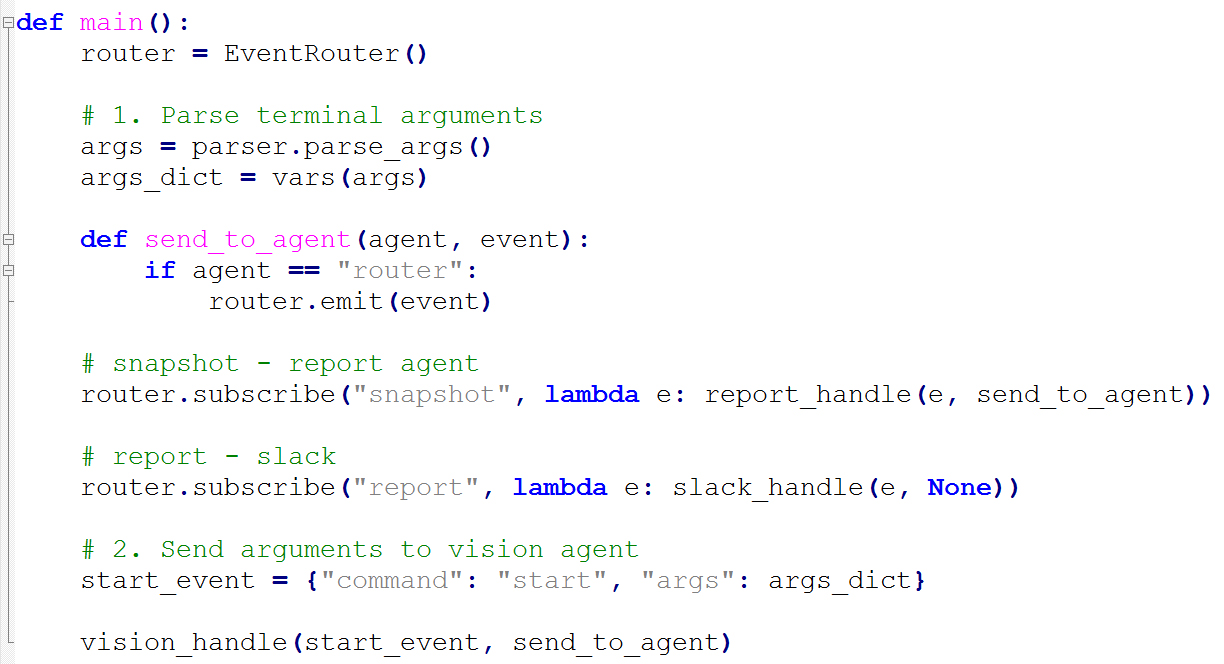}
    \caption{}
    \label{fig: Software components_b}
  \end{subfigure}
  \captionsetup{justification=justified}
  \caption{Software components implementing proposed multi-agent architecture in Fig.~\ref{fig: System architecture}: (a)~modules structure, (b)~code snippet from "main.py" illustrating system initialization phase with reporting and communication agent subscription to agent specific router events.}
  \label{fig: Software components}
\end{figure}

Structure of developed software components implementing proposed multi-agent architecture in Fig.~\ref{fig: System architecture} is shown in Fig.~\ref{fig: Software components_a}, where the corresponding folders or modules are depicted. It is a reminiscent of initial decision to implement communication and control agents as single software component named "slack\_agent", however the final solution , Fig.~\ref{fig: System architecture} clearly differentiates between them.

Fig.~\ref{fig: Software components_b} illustrates the system initialization process, initiated by the user message via Slack channel to the control agent,  Fig.~\ref{fig: System architecture}. It is  implemented by the preloaded system module named "slack\_listener", which is responsible for launching the main initialization script. In this process, reporting and communication agent are subscribing to event driven router messages, Fig.~\ref{fig: Software components_a}, which alongside the launching of the vision agent with parsed command arguments constitutes the main part of the system initialization process.

\subsection{Vision Agent}

Vision agent implements multi-class object detector and accompanying object tracking, which makes it a versatile tool for different application scenarios. Its main characteristic is small memory footprint (in comparison to other system elements, like LLM) and near to real-time operation. As the model of interest for fast prototyping of described multi-agent architecture were chosen single stage detectors from the YOLO family and their highly optimized implementations provided by \cite{Ultralytics2026}. Simultaneous tracking of detected objects was added to enhance system capabilities in possible surveillance and monitoring applications, by providing identification of the same object or higher level inference over multiple frames of the input camera feed. In order to lower the computational cost of the agent operation, tracking was implemented in a passive manner, by utilizing corresponding search algorithm and simple matching criteria based on the intersection between detections' bounding boxes, Algorithm~\ref{alg:tracking}. Predefined setup for vision agent's events generation in analyzed experiments was that the object form the chosen category has been detected and that the same object was present in the scene for the predefined period after initial detection. This trigger corresponds to possible surveillance scenarios in which the system is alarmed after the same object stays for too long in the same zone of the scene. Obviously, there are numerous other event scenarios, but the described simple choice was made primarily to provide an efficient event generation for the analysis of the proposed multi-agent architecture performance on the selected low-resource platform.

\subsection{Reporting Agent}

Events generated by the vision agent are resolved by the reporting agent, Fig.~\ref{fig: System architecture}, which generates event related reports, but by doing so employs a LLM based subsystem. This process is performed asynchronously, as depicted in Algorithm~\ref{alg:reportflow}, where it is shown that the system residing LLM is invoked in a separate thread through the localhost POST HTTP request,  which does not require acknowledgment in the sense of a formal handshake protocol. Report generation time and the report quality depend on the specific model size and performance, respectively. However, the reporting results are also influenced by the prompt formulation provided by the format arguments function of the Algorithm~\ref{alg:reportflow}.

\begin{figure*}[htb]
\centering
\begin{minipage}{0.48\textwidth}
\centering
{\small
\begin{algorithm}[H]
\caption{Passive Tracker Update}
\begin{algorithmic}[1]
    \Procedure{PassiveTrackerUpdate}{$\mathcal{T}$, $\mathcal{D}$, $\theta$, $L_{\max}$}
        \State \textbf{\textit{Init:}} $\mathcal{T}' \gets \emptyset$, $\mathcal{U} \gets \emptyset$, $\mathcal{R} \gets \emptyset$
        \State \textbf{\textit{\#1: Greedy matching}}
        \State \textbf{\textit{For each}} $t_i \in \mathcal{T}$:
        \State \hspace{1em}$j^* \gets \arg\max_{j\notin\mathcal{U}} \text{IoU}(b_i,d_j)$
        \State \hspace{1em}\textbf{\textit{If}} $\text{IoU}(b_i,d_{j^*}) > \theta$:
        \State \hspace{2em}$\mathcal{T}'[i] \gets \{d_{j^*}, l=0\}$
        \State \hspace{2em}$\mathcal{U} \cup=\{j^*\}$, $\mathcal{R} \cup=\{(i,d_{j^*})\}$
        \State \hspace{1em}\textbf{\textit{Else:}} $t_i.l \gets t_i.l + 1$
        \State \hspace{2em}\textbf{\textit{If}} $t_i.l < L_{\max}$: $\mathcal{T}'[i] \gets t_i$
        \State \textbf{\textit{\#2: New tracks}}
        \State \textbf{\textit{For each}} $d_j \notin \mathcal{U}$:
        \State \hspace{1em}$\text{ID}_{\text{new}} \gets \text{ID}_{\text{new}} + 1$
        \State \hspace{1em}$\mathcal{T}'[\text{ID}_{\text{new}}] \gets \{d_j, l=0\}$
        \State $\mathcal{T} \gets \mathcal{T}'$, \textbf{\textit{Return:}} $\mathcal{R}$
    \EndProcedure
\end{algorithmic}
\label{alg:tracking}
\end{algorithm}
}
\end{minipage}
\hfill
\begin{minipage}{0.48\textwidth}
\centering
{\small
\begin{algorithm}[H]
\caption{Reporting Flow}
\begin{algorithmic}[1]
    \State \textbf{\textit{Trigger:}} $event.type=$"snapshot"
    \Statex
    \State \textbf{\textit{Background daemon thread:}}
    \Statex $\downarrow$
    \State \Call{generate\_and\_send}{\emph{event}}
    \State \hspace{1em}$path,detections,timestamp,$
    \Statex \hspace{3em}$args \gets event.payload$
    \Statex $\downarrow$
    \State \Call{format\_args}{\emph{args}} $\rightarrow args\_str$
    \Statex $\downarrow$
    \State \Call{generate\_caption}{\emph{path}, \emph{detections}, \emph{timestamp}, \emph{args\_str}}
    \State \hspace{1em}$\Call{POST}{OLLAMA\_localhost} \rightarrow caption$
    \Statex $\downarrow$
    \State \Call{make\_event}{"report", \{\emph{path},\emph{caption}\}} $\rightarrow report\_event$
    \State $\downarrow$
    \State \Call{send\_to\_agent}{"router",\emph{report\_event}}
\end{algorithmic}
\label{alg:reportflow}
\end{algorithm}
}
\end{minipage}
\caption{Illustration of the implemented functionalities:  (a) vision agent's object tracking, and (b) report agent's asynchronous LLM system call.}
\label{fig: Agent algorithms}
\end{figure*}

\subsection{Communication and Control Agents}
\begin{figure}[htb]
    \centering
    \includegraphics[width=\linewidth]{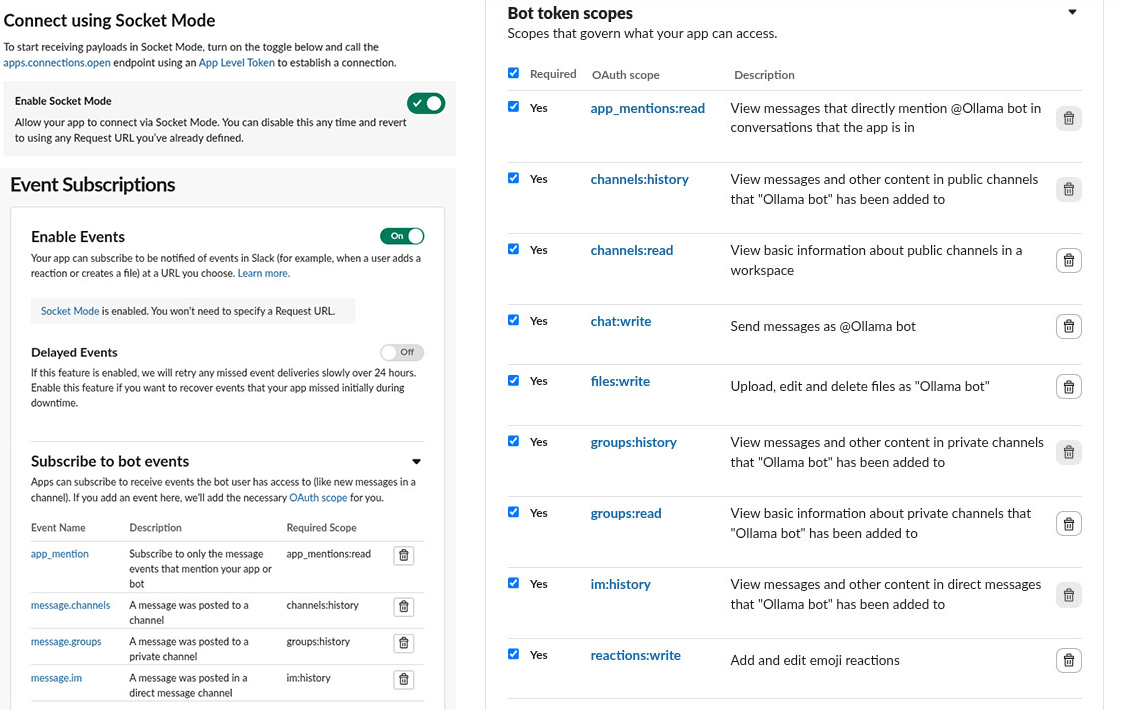}
  \captionsetup{justification=justified}
  \caption{Characteristics of the messaging application: socket mode communication, message events subscription and the bot token scopes defining what functionality communication and control agents can ask for from the chatbot instance.}
  \label{fig: Slack_configuration}
\end{figure}

As illustrated in Fig.~\ref{fig: Software components_a}, communication and control agents have been implemented inside the same software module, the so called "slack\_agent", which incorporates the Slack channel listener (control agent) and communication agent resolving the raised report generation events. Since the used communication infrastructure for the proposed natural language interface assumed the Slack messaging application, in Fig.~\ref{fig: Slack_configuration} are provided some more details regarding its configuration and operation modes. We note that the choice of messaging application is completely independent from the proposed system architecture, and as the result any other messaging provider with chatbot functionality could be applied. The main goal of having this infrastructure was to avoid unnecessary software development of components that would provide graphical interface and type of interaction that users can experience by using mobile or web based messaging clients.

\subsection{Experimental Setup}

Developed prototype of the proposed mult-agent object detection framework was tested in several working modes and system configurations corresponding to potential application scenarios. Namely, the setup investigated differences in prototype operation in cases when all agents were active, against computationally less demanding modes when only a subset of agents was utilized.

The main configuration assumed the full version of the architecture in Fig.~\ref{fig: System architecture} and the reporting agent's LLM subsystem running locally, resulting in peak system load. Alternative configurations considered possible use cases where LLM operation could be off-loaded via an external API, when the system load would mainly be attributed to vision agent's object detection and tracking. In all analyzed working modes the system performance and resource allocation were evaluated based on the reported performance metrics, visual assessment of object detections and tracking quality, reporting speed and  communication channel event responsiveness. Besides varying number of active agents, analyzed working modes mostly differed in the type of utilized object detection models from the YOLO family, since the model size was directly affecting resource allocation of the vision agent's detection subsystem, while the resulting performance of the chosen object detection model influenced passive tracking quality and the vision event reporting frequency. Similarly also holds for LLM subsystem, where models with different number of parameters required different amounts of resources, but also caused different delays and quality of the reporting agent operation. All tests were conducted in real-time without predefined scene dynamics, by placing the prototype in the roles of live public street surveillance from the distance (e.g. by filming cars and pedestrians on the street) or indoor operation (resulting in significantly larger scale of the objects).

\section{Results and Discussion}

\begin{figure}[htb]
    \centering
    \includegraphics[width=0.8\linewidth]{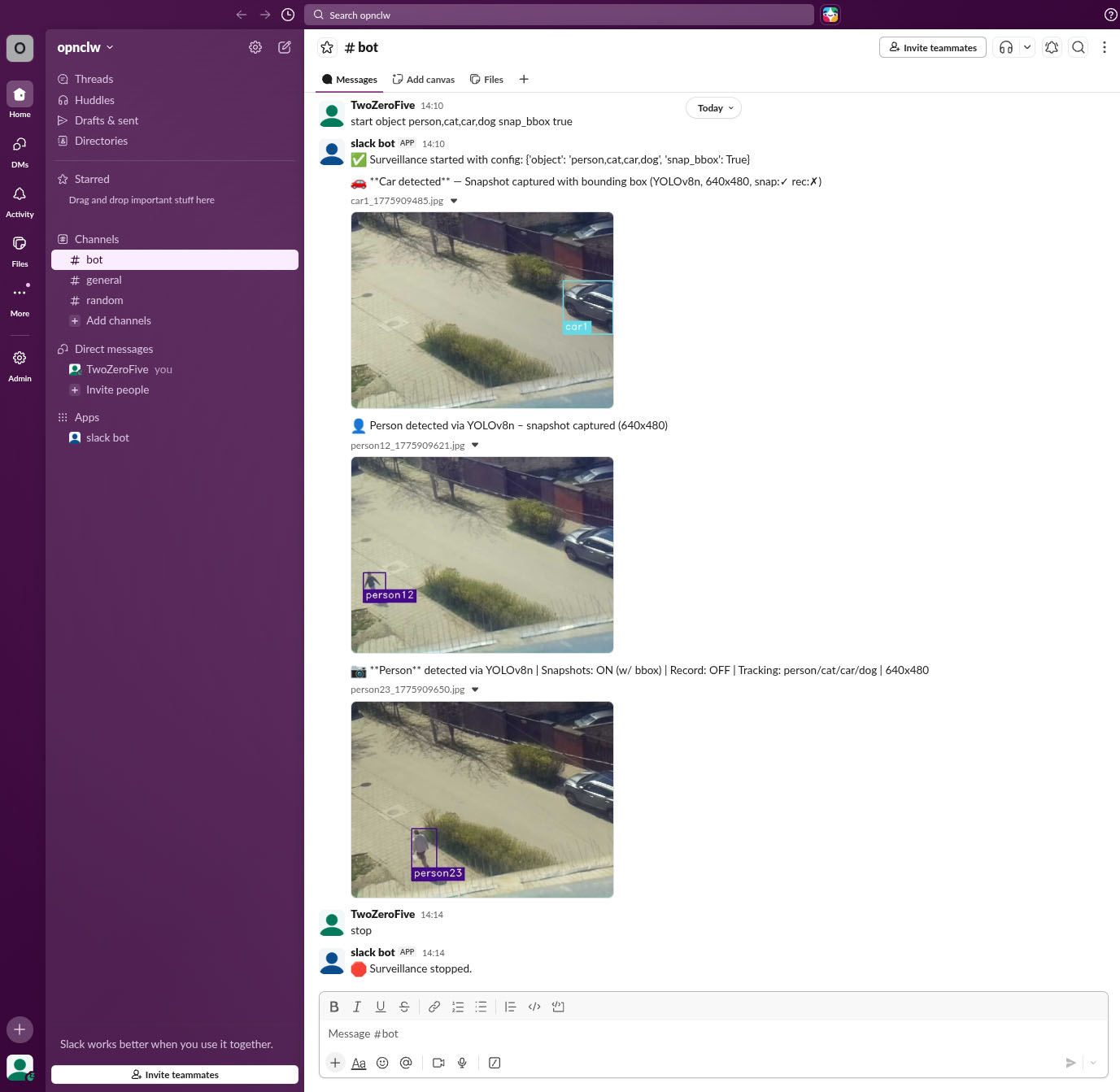}
    \captionsetup{justification=justified}
    \caption{Multi-agent system operation with user interaction through created natural language interface based on the Slack channel and Ollama event report generation. Both notification and control commands are provided through familiar messaging application, while the object detection and tracking events are triggering combined image and text system responses.}
    \label{fig: System operation}
\end{figure}

Described design approach based on the Raspberry Pi platform resulted in a fully functional system, Fig.~\ref{fig: System operation}, implementing the proposed architecture in Fig.~\ref{fig: System architecture}.

Presented solution showcases that the proposed multi-agent object detection framework can be an effective design template for similar edge based AI systems running on the low-cost embedded computing platforms. It also demonstrates that the transformative potential of generative AI systems can significantly reduce development costs of such systems, by leveraging on LLM based communication and control interfaces instead of developing specialized subsystems or infrastructure for the same roles. This holds both for development and deployment phase, where natural language interface can ease system testing, as well as later adoption by the users. Thus, the initial hypothesis that the integration of multiple AI components could be done on the low resource platform in a fully centralized way was confirmed. However, the results of performed experiments also clearly indicate that such integration is still far from being perfect.

Although all agents were capable of performing defined tasks, in experimental configurations involving locally run LLM models, the computational burden was too high, resulting in delayed report generation for system generated vision events. Namely, although the report generation was performed in asynchronous mode, in case of multiple events it resulted in timeouts since the rate of token processing by the locally run LLM was relatively modest. Even in the working regimes with low number of vision events, the time necessary for report generation was at the order of tens of seconds (despite the carefully and conservatively designed LLM input prompts). It means that this type of system configuration with fully centralized operation of all subsystems would not be applicable to scenarios in which reporting event handling and fast user notification would be of utmost importance. However, in cases when certain notification delay would be acceptable, this configuration would be of interest.

Regarding the working modes, the experiments evaluated system performance for several combinations of YOLO based object detection models and Ollama provided LLMs. We have mainly tested the so caled n type of YOLO models provided by \cite{Ultralytics2026}, which are the smallest and the fastest versions of the corresponding YOLO architectures. Thus, besides the latest  yolo26n \cite{yolo26ultralytics}, the following models were also tested: yolo11n \cite{yolo11ultralytics}, yolo12n \cite{yolo12}, yolov5nu \cite{yolov5}, yolo5su, yolov8n \cite{yolov8ultralytics}. In overall, the best performing model for the utilized detection and tracking setup was yolov8n, which produced the most reliable tracking output. Namely, the more complex or accurate models usually come with the price of lower frame rate, resulting in frequent tracker reinitializations, e.g. in the case of fast moving objects like cars or bicycles in the considered street surveillance experiments.

Thus, the vision agent equipped with yolov8n object detector was mostly tested with the provided Ollama models. This included tinyllama:latest (637~MB) \cite{tinyllama}, llama3.2:1b (1.3~GB) \cite{llama3herdmodels}, gemma3:latest (3.3~GB) \cite{gemma2025}, and kimi-k2.5:cloud (non-local) \cite{kimi}, which were selected as the most appropriate (smallest in size), but also with varying memory footprints in order to analyze system performance under different loads. The best performing model, in terms of trade-off between generated report quality and the speed of processing was  llama3.2:1b, however as already mentioned it was not performing at expected speed. It resulted in delayed system notifications, and in the worst cases lost notifications due to report generation timeouts, besides the low frame rate (less then 3 fps at 640x480 spatial resolution of the input image). Nevertheless, it showed that it was possible to achieve full configuration setup with all agents.

\begin{figure}[tb]
  \centering
  \begin{subfigure}{0.35\linewidth}
    \centering
    \includegraphics[width=0.44\linewidth]{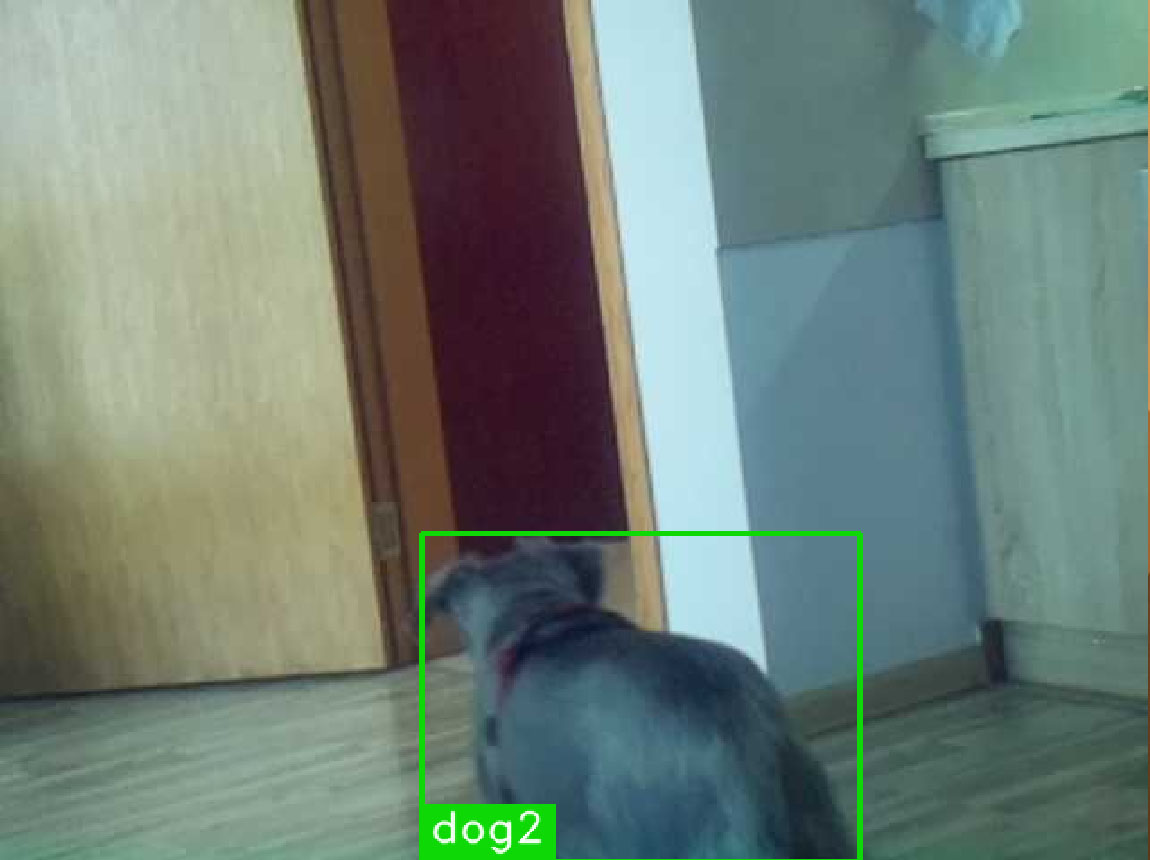}

    \includegraphics[width=0.44\linewidth]{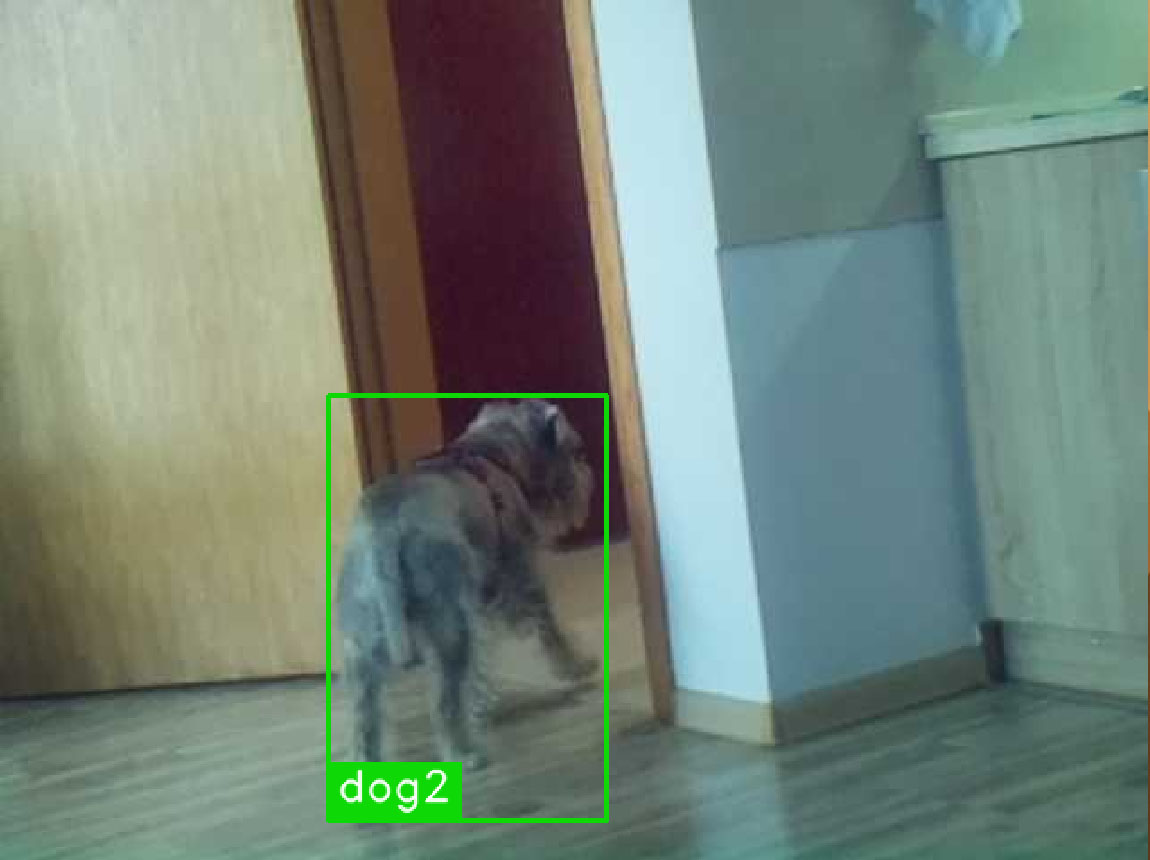}

    \includegraphics[width=0.44\linewidth]{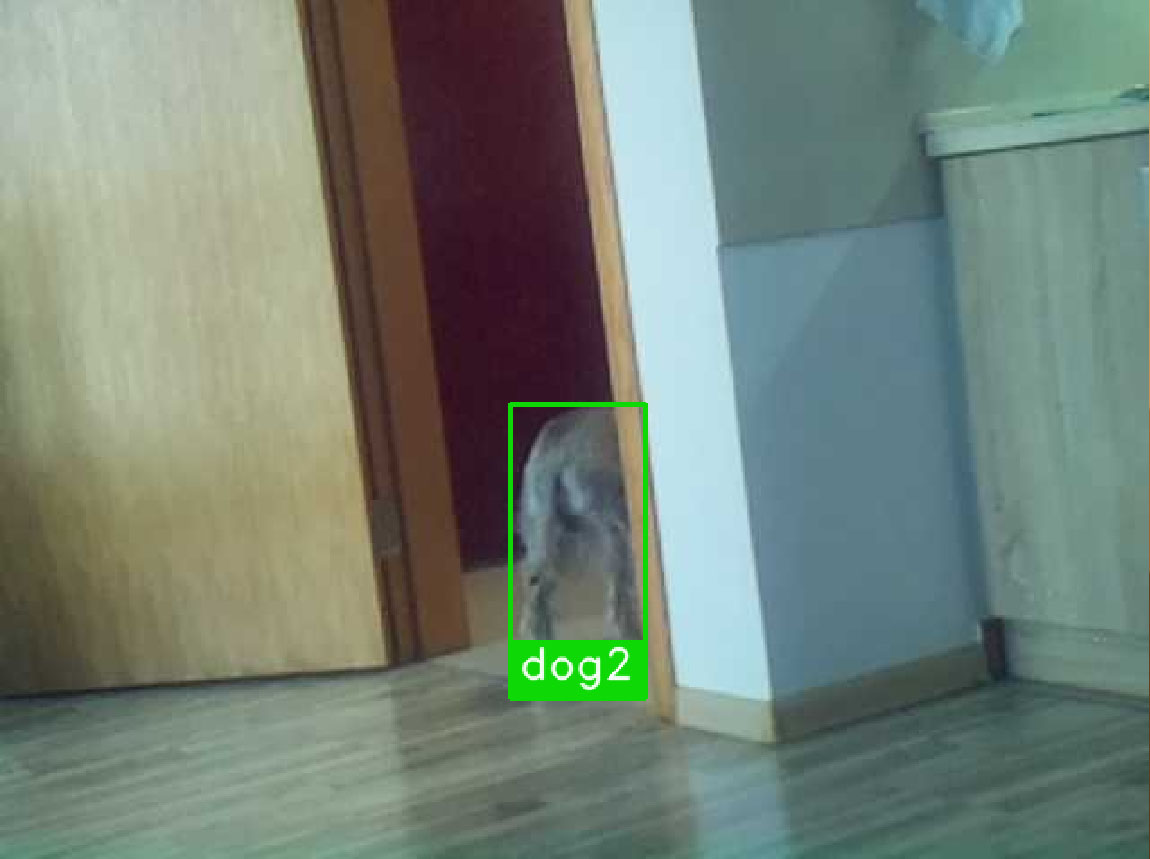}

    \caption{}
    \label{fig: Vision agent performance_a}
  \end{subfigure}
  \hfill
  \begin{subfigure}{0.64\linewidth}
    \centering
    \includegraphics[width=0.85\linewidth]{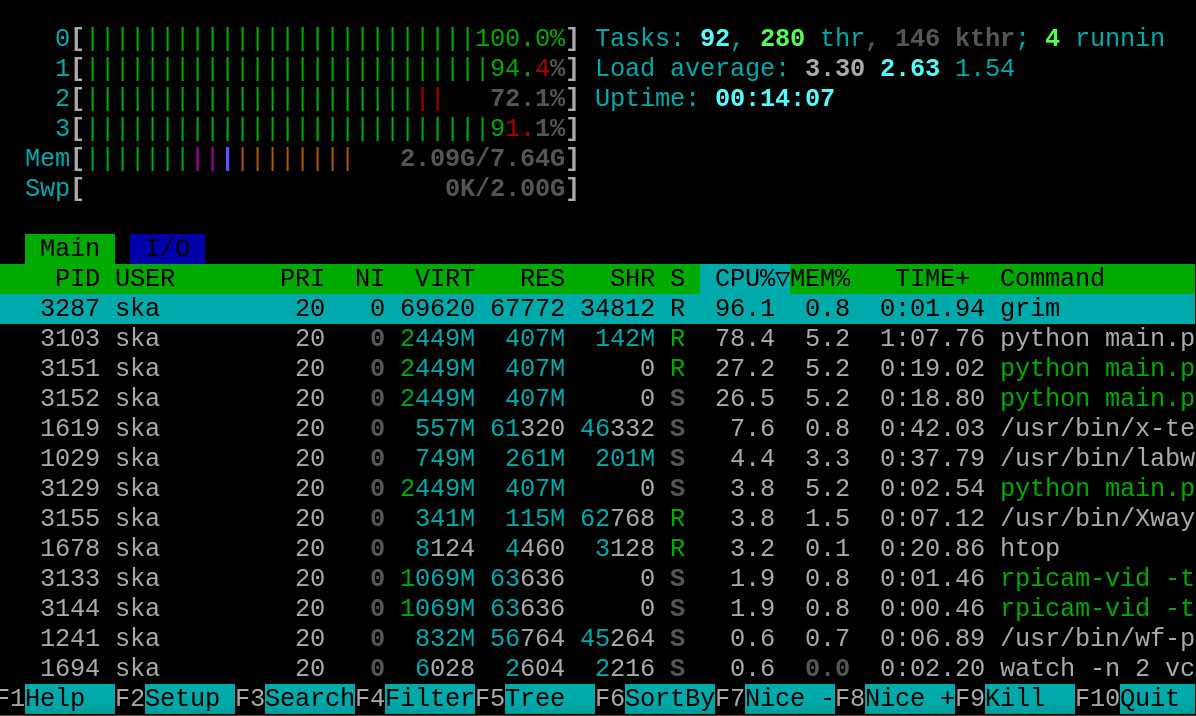}
    \caption{}
    \label{fig: Vision agent performance_b}
  \end{subfigure}
  \captionsetup{justification=justified}
  \caption{Illustration of system configuration without LLM based reporting agent: (a)~object tracking and video encoding, (b)~CPU load and memory usage.}
  \label{fig: Vision agent performance}
\end{figure}

The reason for such performance was probably due to computationally heavy vision tasks that were performed by the vision agent. This can be seen from Fig.~\ref{fig: Vision agent performance}, where it is shown that even in the configurations when only the vision agent was active (without LLM) the quad core CPU load was still very high, Fig.~\ref{fig: Vision agent performance_b}. Besides different working modes, experiments were also done by varying system configuration parameters like object categories, image resolutions, tracking and detection thresholds, live preview or visualization. It was done through the Slack channel interface, allowing for simple reconfiguration and user interaction.

Regarding the initial considerations of using a fully automated and AI based agent orchestration in the proposed solution. The conducted experiments clearly indicated that the OpenClaw design paradigm would require significantly higher token processing speeds by the locally run LLM. This holds primarily for agents orchestration, which was done by the proposed computationally lightweight event driven pipeline, implemented by the router in Fig.~\ref{fig: System architecture}. However, it also refers to the potential automated development of multi-agent systems that would be done by the OpenClaw type application running entirely on the local low-cost platform (by using locally hosted LLM). This finding is also in line with the reported system operation based on the kimi-k2.5:cloud LLM, which has provided the best results in combination with all vision agent models, but required that the reporting agent utilizes LLM subsystem that is not present locally on the Raspberry Pi. Instead, in these alternative system configurations LLM subsystem was running in cloud environment. We note that the integration of the OpenClaw type orchestrator on the selected platform was also tested, but it was not possible to operate such orchestrator without an external cloud based LLM.

When it comes to the market implications of the fast prototyping based on the generative AI, obtained result confirm the ongoing trend of fundamental transformation of the way how information-communication technologies are designed and implemented. Multi-agent systems open possibilities for relatively simple integration of the existing infrastructure, by providing additional value to the previous investments, and making them the basis for new products and services. Therefore, the proposed object detection framework can be also regarded as a template for engineering solutions that avoid expensive and long development cycles or the need for reimplementation of interfaces in order to ensure their integration into new systems and roles.

\section{Conclusion}

Advantages of the proposed low-cost multi-agentic framework are primarily reflecting the design goal of achieving integration of complex AI based system components into compact hardware constrained platform. By demonstrating an implementation of the multi-agent system prototype based on the Raspberry Pi device it was possible to investigate and analyze the overall system performance under the different configuration setups and the working modes. This has revealed practical limitations of the current platform, but also emphasized the modular nature of the proposed architecture and software implementation.

The presented design scheme provides straightforward guidelines for further improvements through implementation of more advanced and customized AI models and accompanying algorithms, or the choice of other edge devices. Manyfold application possibilities also come from the design that supports fast prototyping through simple integration and connectivity, at both internal level between the agents, as well as towards the external environment through the natural language interface.

Compact dimensions, small form factor and low power consumption that were directing the framework design from the start, guarantee that the systems based on the provided template would be simple to use and integrate into larger industrial products or services. However, this does not exclude the framework implementation on a larger and more powerful GPU equipped full scale systems. The advantage of the low-cost edge device implementation is also the expected higher scalability and adaptation to budget constraints when the large number of vision agents would be necessary, i.e. enabling reliable device production. Expected higher market availability and simpler logistics of low-cost devices in the case of bulk procurements would be also an advantage.

Presented results confirmed that it is possible to integrate computer vision and LLM components into the same resource constrianed physical system, but also that this is still challenging since the current solutions are still not compact enough to provide desired operation (e.g. without notification delays in the case of LLM based report generation). However, this does not mean that the locally run LLM instances, like Ollama models, are not possible to fully operate on the hardware constrained platform. Actually, conducted experiments involving only LLM based agents confirmed that they can be fully functional part of the system and easily accessed through the designed Slack channel interface. Thus, these properties make them a good fit for application scenarios in which real-time responsiveness is not a mandatory requirement. However, when other computational heavy AI agents are also present in the system, advanced NLP functionalities are hard to achieve without an external LLM inference. Similarly also holds for the recently proposed OpenClaw AI agent orchestration paradigm, which was possible to implement on the selected platform, but was not possible to utilize without an external LLM instance.

In the future work we plan to further address these challenges and explore more specific use case scenarios involving similar edge devices.

\begin{credits}
\subsubsection{\ackname} The second author would like to acknowledge support by the Science Fund of the Republic of Serbia through the project AI-SPEAK - "Multimodal multilingual human-machine speech communication", grant agreement no. 7449;  and the Ministry of Science, Technological Development and Innovation of the Republic of Serbia (Contract No. 451-03-34/2026-03/200156), for the support through the project “Scientific and artistic research work of researchers in teaching and associate positions at the Faculty of Technical Sciences, University of Novi Sad 2026”, No. 01-3609/1.

\subsubsection{\discintname} The authors have no competing interests to declare that are
relevant to the content of this article.
\end{credits}

%
%

\bibliographystyle{IEEEtranDOI}
\bibliography{SYMORG2026}

\end{document}